\documentclass{article}
\usepackage{spconf,amsmath,graphicx}
\usepackage{diagbox}
\usepackage{url}
\usepackage{graphics}
\usepackage{calc}
\usepackage{setspace}
\usepackage{microtype}
\usepackage[absolute]{textpos}
\newcommand{\myfigureshrinker}{\vspace{-1mm}}
\TPMargin{5pt}

\newcommand{\copyrightstatement}{
    \begin{textblock}{0.83}(0.09,0.92)    % tweak here: {box width}(leftposition, rightposition)
		\tekstblokkulur{white}%         
         \noindent
         \scriptsize
         \copyright \, 2021 IEEE. Personal use of this material is permitted. Permission from IEEE must be obtained for all other uses, in any current or future media, including reprinting/republishing this material for advertising or promotional purposes, creating new collective works, for resale or redistribution to servers or lists, or reuse of any copyrighted component of this work in other works.
    \end{textblock}
}

\title{ZERO IN ON SHAPE: A GENERIC 2D-3D INSTANCE SIMILARITY METRIC LEARNED FROM SYNTHETIC DATA}
\name{Maciej Janik, Niklas Gard, Anna Hilsmann, Peter Eisert
}
\address{Fraunhofer HHI, Berlin, Germany}
\begin{document}
\maketitle
\begin{abstract}
\vspace{.1em}
We present a network architecture which compares RGB images and untextured 3D models by the similarity of the represented shape. Our system is optimised for zero-shot retrieval, meaning it can recognise shapes never shown in training. We use a view-based shape descriptor and a siamese network to learn object geometry from pairs of 3D models and 2D images. Due to scarcity of datasets with exact photograph-mesh correspondences, we train our network with only synthetic data. Our experiments investigate the effect of different qualities and quantities of training data on retrieval accuracy and present insights from bridging the domain gap. We show that increasing the variety of synthetic data improves retrieval accuracy and that our system's  performance in zero-shot mode can match that of the instance-aware mode, as far as narrowing down the search to the top 10\% of objects.

\end{abstract}
\copyrightstatement
\section{Introduction}
\label{sec:intro}
Reasoning about real objects in the context of computer vision is easier if we are in possession of their 3D models. Unlike 2D images, 3D models are not spoiled by factors such as viewpoint, occlusion, or lighting, and therefore provide a pure and accurate reference to the real object. In particular, we are interested in rigid objects which can be reliably recognised based solely on their geometry.

\begin{figure}
\myfigureshrinker
  \includegraphics[width=\linewidth]{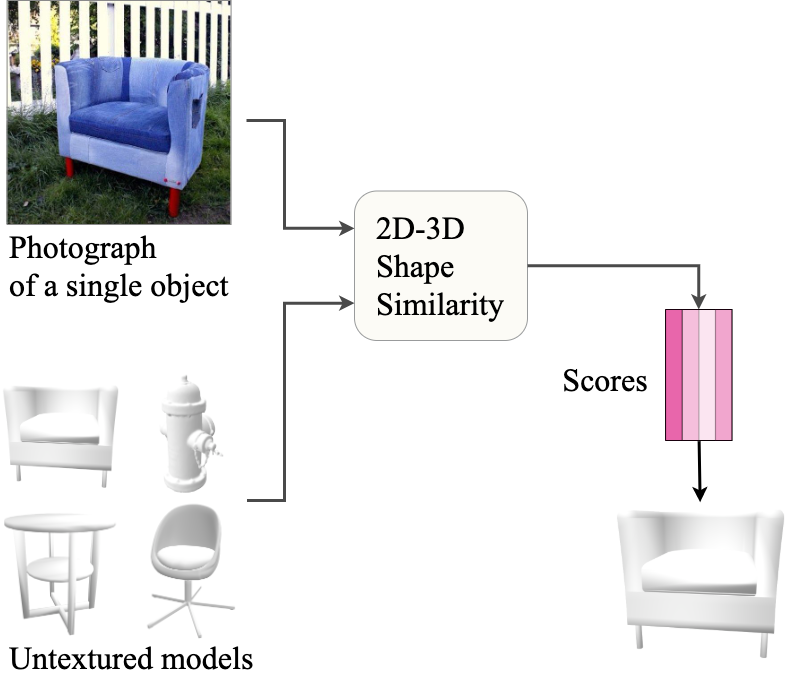}
  \caption{Our network retrieves an unknown 3D model from an RGB image on the sole basis of object geometry.}
  \label{fig:metric}
\end{figure}

In this paper, we learn a metric (Figure \ref{fig:metric}) which measures shape similarity of objects in RGB images and as 3D models. The metric is implemented as a Siamese CNN \cite{Bromley93} which extracts shape features from images and models and, for identical shapes, maps them close together in the latent space. 
We represent 3D models by a set of textureless, shaded greyscale views, rendered from viewing positions on the corners of an icosahedron enclosing the object. 
Features computed from every view are merged into a compact representation by max-pooling, similar to \cite{Su15}. 
Similarity was learned by exposing the CNN to same and different image-model pairs, computing cosine distance between their extracted features and applying the contrastive loss function \cite{Hadsell06}.

In our work, we focus on zero-shot retrieval \cite{zero_shot}, that is, on optimizing the network to recognise shapes different from those shown in training. This ability to work with arbitrary shapes makes our network very practical, since making changes (such as adding new objects) to the retrieval set does not require any retraining of the model.
To this end, we expose the CNN to a sufficient number of versatile image-to-model correspondences such that it learns a generic 2D-to-3D shape similarity function.

Due to scarcity of existing datasets featuring real images and 3D models of the exact same objects, we train our network using only synthetic data. We generate the data using Domain Randomization \cite{Tobin17} and leverage the siamese architecture of the CNN to reduce the domain gap between the synthetic training data and the real test data.

In summary, our contributions are three-fold: 1) we design a network which finds a common embedding of two heterogeneous object representations 2) we demonstrate the network's potential for recognising unknown objects (zero-shot) 3) we present insights from training the network with only synthetic data and from bridging the domain gap.

\section{Related Work}
\label{sec:state}

\paragraph*{Image-Based 3D Model Retrieval}
\label{sec:retrieval}
In order to match images with 3D models we must first design suitable embeddings of these media. Compared to image descriptors, shape descriptors are still relatively unexplored. While shape descriptors can be computed from native model formats, such as a polygon mesh or a point cloud \cite{Feng19, Pham19}, most of the existing approaches instead employ intermediate representations of models, such as grayscale renderings \cite{Su15}, depth maps \cite{Grabner18}, or location fields \cite{Grabner19}. 
Such descriptors were shown to perform better in both model-to-model and model-to-image comparison tasks \cite{Su15}.
Besides, operating on 2D projections, rather than on native 3D formats, is computationally more efficient and allows leveraging existing image descriptors pretrained on large datasets such as ImageNet \cite{imagenet}.

Another question is how to jointly embed shapes and images in a meaningful way. 
Our work is most similar to the one of \cite{Lee18} who also aggregate greyscale views of each model into a single vector, and map them together with RGB images with a siamese network. Different from us, they train their network using real images and corresponding shape instances from 40 different classes. As their training pairs use the closest available shape, rather than the exact same shape, their system effectively works on a class-level, rather than on the instance-level.

Grabner et al.~propose to match images and shapes by computing location field descriptors from both media \cite{Grabner19}. The authors train their network for instance-level recognition and achieve state of the art results on the Pix3D dataset \cite{pix3d}. However, the small size of the dataset restricts them to learning only a small number of shape classes. In contrast, in our work we extend the recognition to unknown classes. For this we employ (unlimited in samples) synthetic data, which shifts a portion of our work towards bridging the synthetic-to-real domain gap.

\paragraph*{Synthetic Data}
\label{sec:synthetic}
In recent years, visual deep learning saw a growing trend for using synthetic images \cite{Nvidia18, Tremblay18, Hinter19}, driven by the scarcity of available real data.
Image synthesis allows to quickly generate fully annotated images in potentially unlimited quantities. Recently emerged tools such as BlenderProc \cite{blender} and NDDS \cite{ndds} provide ready-made dataset generation pipelines and example maps for rendering custom objects. On the downside, synthesized images have different characteristics from the real ones. This results in a 'domain gap' -- systems trained with synthetic data to show worse results on real data than if they had been trained with real data.

Two of the common approaches to mitigating this problem are Domain Randomization (DR) and Domain Adaptation (DA). DR, introduced by Tobin et al.~\cite{Tobin17} for training an object detector, proposes to augment the simulated scene in a large number of ways, such as by randomising object poses, textures, camera poses, and lighting, intending that the real scene would appear to the network just as one of the many synthetic variations. 

Although most works use synthetic data only for pretraining or complementing real data, Tremblay et al.~\cite{Tremblay18} attempted to bridge the domain gap without using any real data. Their training set featured 50\% non-photorealistic DR renderings and 50\% photorealistic renderings. The authors showed that mixing these two types of data results in better performance then using either of the data types alone. Their all-synthetic dataset was used for 6D pose estimation of 21 textured objects, and it has shown competitive results against detectors trained with both real and synthetic data.

While DR is applied to the generation of data, DA aims to make up for different distributions of computed image features, usually in the direction from real to synthetic. For example, Massa et al.~\cite{Massa16} add to their system an adaptation layer pretrained to map synthetic images of certain objects to their real counterparts. Lee et al.~\cite{Lee18} adopt a similar approach, but their adaptation function is learned together with the rest of the CNN in an end-to-end manner.

Our approach follows the DR principle, such that it varies all parameters of the simulated scene except for the 3D shape of interest rendered in the foreground. This prompts the CNN to extract the pure shape information from complex real images. We follow Tremblay et al.~\cite{Tremblay18} in using only synthetic training data, as well as making our training set half-photorealistic and half-randomised. Unlike them, we do not restrict ourselves to classifying a small set of textured shapes, but aim for learning a general idea of shape. On top of that we introduce a novel adaptation strategy of splitting the siamese networks in the few beginning layers allowing the CNN to extract the shape information differently from different types of data.

\section{Method}
\label{sec:method}

\begin{figure} 
\myfigureshrinker
  \includegraphics[width=\linewidth]{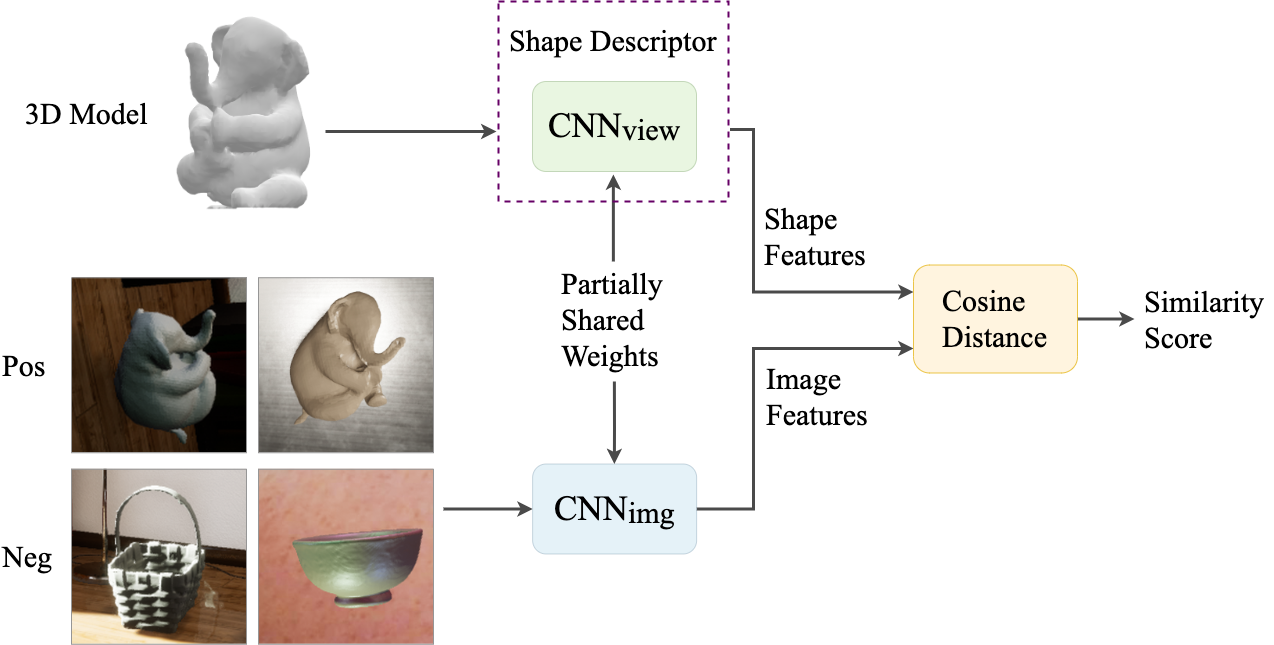}
  \caption{Network Architecture}
  \label{fig:architecture}
\end{figure}

\begin{figure}
\myfigureshrinker
  \includegraphics[width=\linewidth]{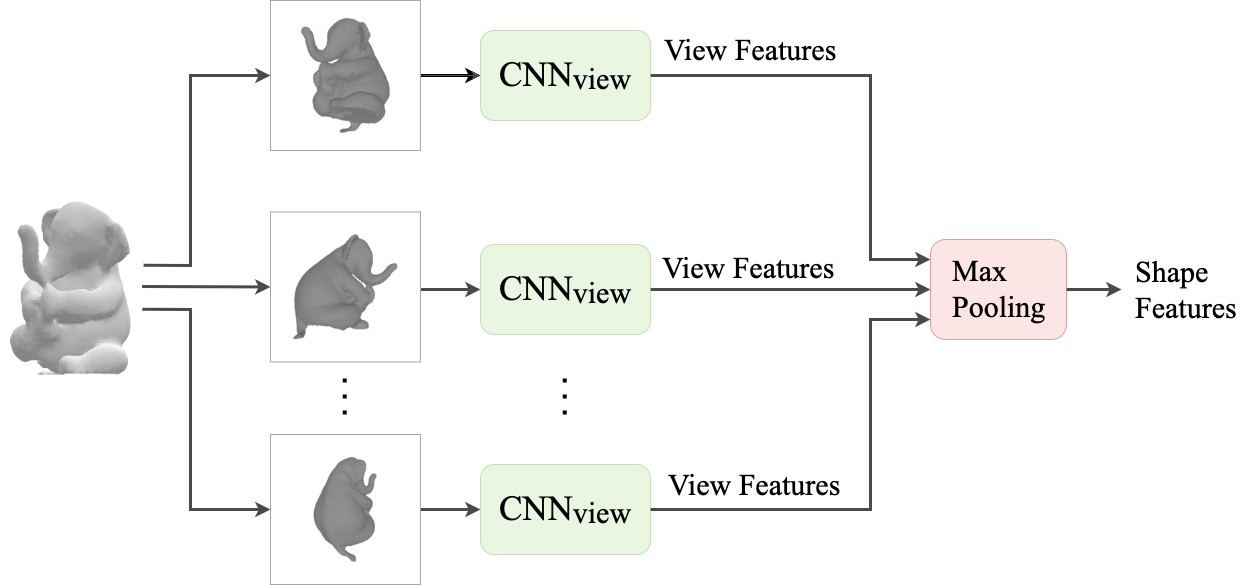}
  \caption{Shape Descriptor}
  \label{fig:view_pool}
\end{figure}

\paragraph*{Network Architecture}
Figure \ref{fig:architecture} presents our network architecture. In each training or evaluation step the network compares one untextured 3D model with one colour image. We use ResNet-34 \cite{resnet} to compute the image descriptors, and reduce the number output features from 1000 to 128. Shape features are computed from twelve all-around views. As shown in Figure \ref{fig:view_pool}, the same \textit{CNN-view} computes features for all views, which are then merged into a single shape representation via max-pooling. In parallel, the \textit{CNN-img} computes features of the colour image (either synthetic or real). \textit{CNN-view} and \textit{CNN-img} share the parameters from the ResNet's 7th layer on; this allows to better adjust to pixel-level differences, while still jointly extracting the relevant shape information in the later layers. Finally the shape vector and the image vector are L2-normalized and the cosine distance computes their similarity score. Based on the pair label and the similarity score, we compute contrastive loss \cite{Hadsell06}, whose margin we set to 1.0.

The training sets were designed such that each object was represented by a roughly equal number of colour images. While the \textit{CNN-view} processed 12 views of an object, the \textit{CNN-img} processed 6 positive and 6 negative images. On average, in one epoch every colour image was fed to the network twice, once as a positive and once as a negative example. In addition, the colour images were augmented with vertical and horizontal flip, random translation, rotation, and scale, as well as with varying contrast and brightness.

\paragraph*{Data Generation}

Experimental data was made up of 3 types of images, all depicting a single object: 1) plain, greyscale renderings, 2) colour renderings, and 3) real photographs. Greyscale renderings, or views, (part of Figure \ref{fig:view_pool}) were generated using Pyrender \cite{pyrender}. Each mesh was given the default metallic material parameters and was placed at the centre of an icosahedron. The camera was placed at each of the 12 vertices and oriented at the object centre. Each captured view was cropped such that the object's longer dimension occupied 70\% of the image. In addition, we tried representing 3D objects with sets of 20 and 42 views, but noted no advantage compared to using just 12.

\begin{figure}
\myfigureshrinker
\centering
  \includegraphics[width=.9\linewidth]{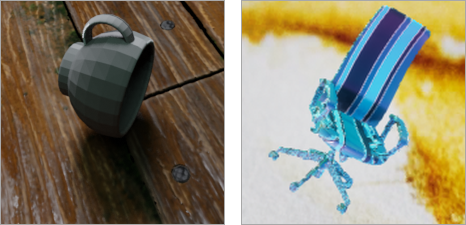}
  \caption{Photorealistic renderings (BlenderProc, left) draw a direct link with the real world, while non-realistic ones (NDDS, right) provide more background and texture variation.}
  \label{fig:renderings}
\end{figure}

Colour renderings served to imitate real images and were generated using BlenderProc and NDDS. One example of each is shown in Figure \ref{fig:renderings}. In BlenderProc, we focused on photorealism of the images. The objects were rendered inside a cube, whose inner walls were assigned varying textures \cite{cctextures}, and where the illumination was frequently randomised. Objects were also assigned varying textures and placed on the floor in randomly sampled poses. Renderings were captured from an upper half sphere centred on the object, with a varying radius. 

In NDDS, the objects were captured in a fully-randomised scene, by rendering objects in varying poses on many different backgrounds. In this case, the realistic look was traded for more variation in object textures \cite{dtd}, backgrounds \cite{flickr} and lighting, the simplicity of the scene and the speed of rendering.

The last type of images are photographs which serve as test data only. The test set features 325 images picked from two datasets: Pix3D \cite{pix3d} and Toyota-Light (a subset of BOP \cite{bop}), and in small part found on web search engines. Each image is annotated with the name of the object it depicts. In order to assure that the objects are well visible, the images were cropped accordingly. We retrieve from a set of 50 models of distinct shapes representing furniture and other household objects. A link to the test set is given in the supplementary material.

\section{Experiments}
\label{sec:exp}

The model was trained in two different modes, which will be referred to as instance-aware and zero-shot. In the first mode the same 50 object instances were used in training and in evaluation. Effectively, the network's task was to transfer knowledge about a set of shapes from synthetic to real data in the most efficient manner. 

In zero-shot mode, none of the test objects were shown to the network in training. Instead, the model learned from data rendered from several hundreds of different 3D models, in pursuit of generalising to recognition of any shape.

The network was implemented in Pytorch \cite{pytorch}. After experimenting with AlexNet \cite{alexnet}, VGG-16 \cite{vgg} and different depth-versions of ResNet, we chose ResNet-34 as the best backbone for our task. We used Adam optimizer \cite{adam}, set the learning rate to 5e-5, weight decay to 1e-5 and the batch size to 12. After every epoch we evaluated the model on the real test set and reported results from the checkpoint with the highest retrieval rate of the single correct object (Top-1 accuracy). Depending on the dataset used this proceeding required training for 10 to 25 epochs.

The initial experiments were conducted in instance-aware mode. Firstly, we investigated how well different types of synthetic data serve learning our function. For this, we rendered two training sets of 6k images, featuring 50 object instances: one using BlenderProc, with the focus on photorealism and one using NDDS, with the focus on more randomisation. Examples of these images are shown in Figure \ref{fig:renderings}. The network was trained thrice: first with only realistic data, then with only randomised data, finally with the union of the datasets. The top accuracy for all trainings are shown in Table \ref{tab:data_types}. We see that the photo-realistic data alone yielded better results than the randomised data alone. Mixing the datasets resulted in the best accuracies of all, which speaks for increasing the variety of synthetic data to train with. 

\begin{table}
\myfigureshrinker
\centering
\resizebox{\columnwidth*\real{0.9}}{!}{%
\begin{tabular}{|c|*{4}{c|}}
\hline
\diagbox{Acc.}{Mode} & Photorealistic & Randomised & Mixed \\ \hline
Top-1 & .50 & .46 & .59 \\\hline
Top-2 & .65 & .61 & .73 \\\hline
Top-5 & .86 & .80 & .87 \\\hline
\end{tabular}%
}
\caption{Comparison of different synthetic training data types.}
\label{tab:data_types}
\end{table}

Secondly, we looked into the effect of siamese parameter sharing between the network's arms. The model was trained once more in instance-aware mode, this time using more training samples (20k) including renderings from two extra UE4 maps, in order to further increase the retrieval rates through more varied data. Following that, the model was retrained with the same data, this time with \textit{CNN-view} and \textit{CNN-img} learning separately. The retrieval rates shown in Table \ref{tab:siamese} demonstrate a clear advantage from sharing weights.

\newcolumntype{P}[1]{>{\centering\arraybackslash}p{#1}}
\begin{table}
\myfigureshrinker
\centering
\resizebox{\columnwidth*\real{0.9}}{!}{%
% \begin{tabular}{|l|*{3}{c|}}
\begin{tabular}{|P{2.12cm}|P{2.50cm}|P{2.50cm}|}
% \begin{tabular}{p{0.1\textwidth}>{\centering}p{0.15\textwidth}>{\centering\arraybackslash}p{0.15\textwidth}}
\hline
\diagbox{Acc.}{Mode} & Separate Params & Shared Params \\\hline
Top-1 & .55 & .62 \\\hline
Top-2 & .71 & .75 \\\hline
Top-5 & .87 & .90 \\\hline
\end{tabular}%
}
\caption{The effect of siamese weight sharing.}
\label{tab:siamese}
\end{table}

\begin{figure}
\myfigureshrinker
  \includegraphics[width=\linewidth]{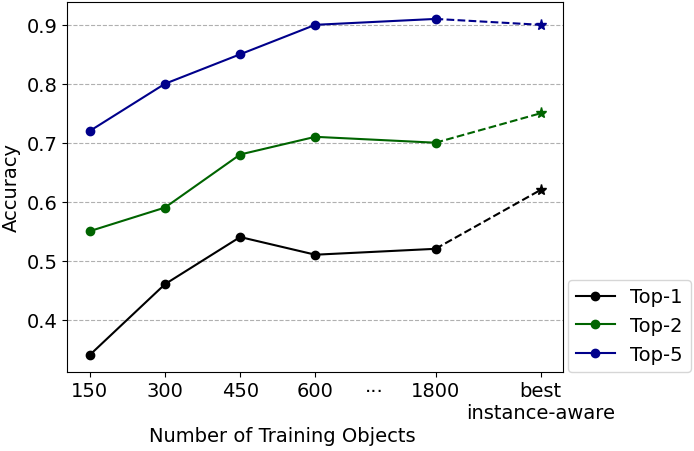}
  \caption{Deriving a sufficient number of objects to learn generic shape similarity. For comparison we also include the retrieval rates of the best instance-aware model.}
  \label{fig:zero_shot}
\end{figure}

The final experiments were conducted in zero-shot mode, as we looked into how increasing the number of training objects affects the retrieval rates. For this the model was trained five times, using sets of 150, 300, 450, 600, and 1800 different meshes sampled from ShapeNet \cite{shapenet}. The best accuracy is shown in Figure \ref{fig:zero_shot}, together with the best accuracy from the instance-aware mode (all models were evaluated on the same test set).
The graph shows an increase in accuracy caused by raising the number of objects, until using 600 unique objects. Further increase in this number does not show a clear advantage. 
If we couple these results with those of Experiment 1, we see that for zero-shot object recognition in the wild it is more vital to increase the versatility of synthetic training data than to increase the number of samples beyond a certain point.

A comparison of the best instance-aware model and the zero-shot model shows that although the former's Top-1 and Top-2 accuracy are clearly better, zero-shot matches the other's performance with the Top-5 accuracy. This enables the use of our metric in applications that only require narrowing down the search to a few candidates.

In a qualitative analysis of retrieval cases we noted that the network tends to have trouble classifying images with strong patterns present in the background as well as those with non-uniform texturing of the object itself.

\section{Conclusions}
\label{sec:conclusions}
We presented a novel approach to matching RGB images and 3D models of arbitrary objects. We showed that our network, trained with only 600 unique objects, can extend its recognition ability from seen to unseen shapes as far as the top five retrieved results are concerned. This zero-shot functionality has practical use in situations where we need to recognise obscure shapes, for which little training data exists or wherever we frequently add new objects to the retrieval set.
Besides, we showed the advantage of siamese weight sharing for learning cross-dimensional similarity of shape, and of increasing the variety of synthetic training data. 

In the future, we plan to explore different shape descriptors, such as MeshNet \cite{Feng19}, as well as to backtrack network activations in order to highlight image pixels most responsible for the extraction of similar features from the two inputs \cite{Seibold2021}.

\section{Acknowledgement}
\label{sec:acknowledgement}
This work is supported by the German Federal Ministry of Economic Affairs and Energy
(DigitalTWIN, grant no. 01MD18008B) and the German Federal Ministry of Education and Research (Single Sensor 3D++, grant no. 03ZZ0457).

%\vfill
%\pagebreak

\begin{spacing}{0.9}
\bibliographystyle{IEEEbib}
\bibliography{strings,refs}
\end{spacing}

\end{document}